\documentclass[10pt,twocolumn,letterpaper]{article}

\usepackage{cvpr}
\usepackage{times}
\usepackage{epsfig}
\usepackage{graphicx}
\usepackage{amsmath}
\usepackage{amssymb}

\usepackage{tabularx}
\usepackage{makecell}
\usepackage{multirow}
\usepackage{subfig}
\usepackage{tabulary}
\usepackage{booktabs}

\usepackage[noend]{algpseudocode}
\usepackage{algorithmicx,algorithm}

\usepackage[breaklinks=true,bookmarks=false]{hyperref}

\cvprfinalcopy 

\begin{document}

\title{Sampling Techniques for Large-Scale Object Detection\\from Sparsely Annotated Objects}
\author{
Yusuke Niitani \,\,\,\, 
Takuya Akiba \,\,\,\,\,\,
Tommi Kerola \,\,\,\, Toru Ogawa \,\,\,\, Shotaro Sano \,\,\,\, Shuji Suzuki \\
Preferred Networks, Inc.\\
{\tt\small \{niitani,akiba,tommi,ogawa,sano,ssuzuki\}@preferred.jp}
}

\maketitle

\begin{abstract}
Efficient and reliable methods for training of object detectors 
are in higher demand than ever, and more and more data relevant to the field is becoming available. However, large datasets like Open Images Dataset v4 (OID) are sparsely annotated, and some measure must be taken in order to ensure the training of a reliable detector. In order to take the incompleteness of these datasets into account, one possibility is to use pretrained models to detect the presence of the unverified objects. However, the performance of such a strategy depends largely on the power of the pretrained model. 
In this study, we propose \textit{part-aware sampling}, a method that uses human intuition for the hierarchical relation between objects. In terse terms, our method works by making assumptions like ``a bounding box for a car should contain a bounding box for a tire''. We demonstrate the power of our method on OID  and compare the performance against a method based on a pretrained model. Our method also won the first and second place on the public and private test sets of the Google AI Open Images Competition 2018.

\end{abstract}

\vspace{-0.5cm}

\section{Introduction}
With recent advances in automation technologies that are dependent on the method of extracting information from the images, the task of object detection has been becoming increasingly important in the field of artificial intelligence. 

Also growing with the interest for the methods of object detection is the size of the dataset that is available for training purpose. Recently published Open Images Dataset v4 (OID) features up to $500$ categories and $1.6$M images with $14$M objects to be detected \cite{kuznetsova2018openimages}.   
It is a dataset on an unprecedented scale in terms of the number of 
annotated images, and each image in the dataset on average contains $\sim7$ categories that were verified by humans (\textit{verified categories}) .    
As a human-annotated dataset, however, the completeness of OID is inevitably somewhat questionable. As claimed in their work, the annotation recall is $43\%$, which means that more than half the objects in the images are missing annotations.

The major problem with this kind of a dataset is that a network would suffer from incorrect training signals due to objects missing annotations.
One naive but sure way to deal with such case is to simply exclude regions surrounding objects missing annotations during the evaluation of the objective function. 
This task, however, is easier said than done, because the validity of this approach depends on our ability to detect the unverified object that is actually present.
One option is to train a pretrained model and use it as an oracle to  tell the presence of unverified object ~\cite{yan2017ssl}.  
Nevertheless, the performance of such an approach is limited by the power of the pretrained model. 

A more intuitive approach is \textit{to utilize the intuition that we are born with}.  If "car" is present in the image,  "tire" should be present in the bounding box of "car" with high probability. If "door" is present in the image, "door handle" should be present in the bounding box of "door" as well. Therefore, leaving some pathological cases side,  if "car" is verified but "tire" is not, then it probably represents the \textit{dangerous case} that we are concerned with---that is, "tire" is an unverified but present object in the image, and we should not include the cost regarding the "tire" in the objective function.  Better yet,  if "car" is present, then we should simply not question the presence of an object of a  \textit{part category} like "tire" in the bounding box of "car".  

This is in fact exactly what our \textit{part-aware} sampling method does.
Given a verified object in the image and a set of all sub-bounding boxes contained in the bounding box of the object, we refrain from asking the detector to detect in the sub-boxes the objects that are, based on our intuition, part of the object captured by the parent bounding box.  

In order to find the better way to detect unannotated objects, we compared our approach against the method based on the pretrained model.
The method using a pretrained model is optimized to work well for sparsely annotated dataset by us.
We call the method \textit{pseudo label-guided} sampling, and verified its effectiveness on OID and artificially created sparsely annotated data based on MS COCO~\cite{yi2014coco}.

Our part-aware sampling leads to an average $0.7$ AP improvement across all categories and $9.2$ AP improvement on \textit{part categories}.
In particular, for human part categories like \textit{human ear} and \textit{human hand}, we confirmed an improvement of $22.7$ AP on average.
Based on our part-aware sampling, we achieved the 1st and 2nd places on the public and private test sets of Google AI Open Images Competition 2018.

\begin{figure}
\captionsetup[subfigure]{labelformat=empty} 
\subfloat[]{
	\label{subfig:unverified}
	\includegraphics[width=0.48\textwidth]{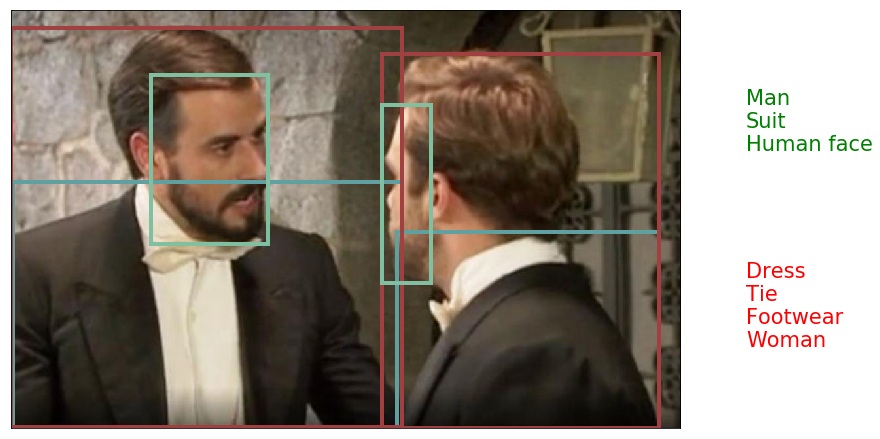} } 
\\
\subfloat[]{
	\label{subfig:verified}
	\includegraphics[width=0.48\textwidth]{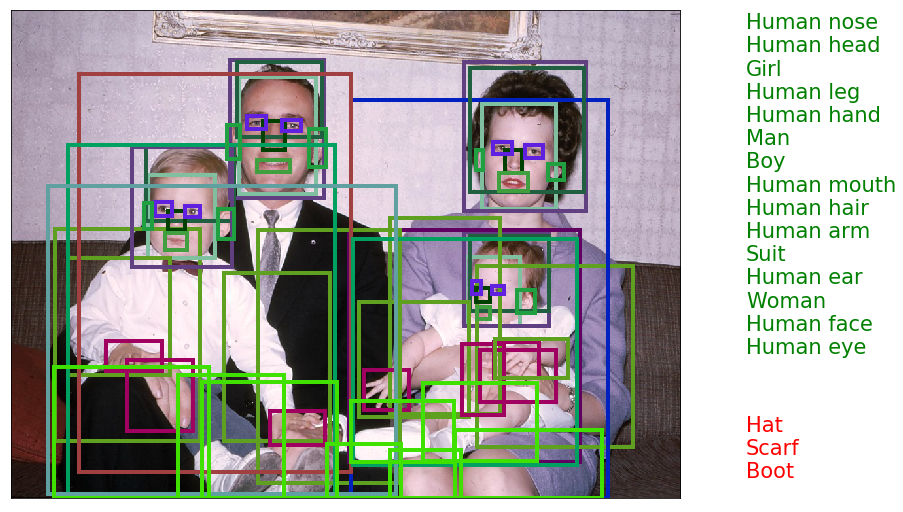} } 
\caption{Example annotations in Open Images Dataset v4. The positively and negatively verified labels are displayed on the right of the images in green and red, respectively.
Many human part categories are absent in the verified labels of the top image.
Such missing annotations create false training signals for a normal object detector.
} 
\label{fig:sparse_verified}
\end{figure}

\begin{figure*}
\subfloat[Human annotations on "human" and "car".]{
	\label{subfig:gt_anno}
	\includegraphics[width=0.30\textwidth]{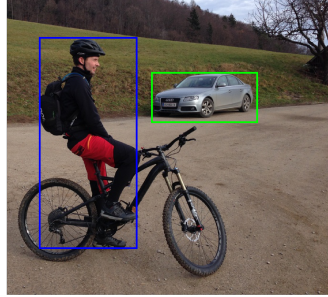} }
\hfill
\subfloat[RoI proposals that are used for training.]{
	\label{subfig:rois}
	\includegraphics[width=0.30\textwidth]{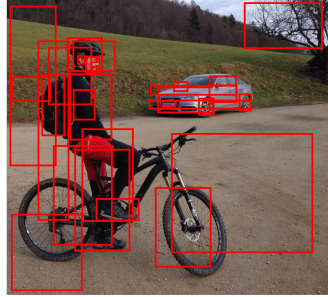} }
\hfill
\subfloat[Blue: Proposals that ignore parts of "person", such as "footwear" and "human face". Green: Proposals that ignore parts of "cars", such as "license plate" and "tire".]{
	\label{subfig:pas_proposals}
	\includegraphics[width=0.30	\textwidth]{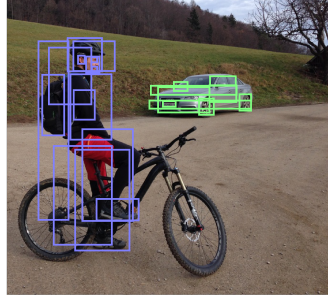} } 
\caption{Description of \textit{part-aware} sampling.
In the left and the middle images, the ground truths and RoI proposals are displayed.
These are the inputs to the algorithm. On the right, we display a subset of the RoI proposals that are ignored for classification loss of certain part categories based on part and subject relationships.}
\label{fig:pas}
\end{figure*}

\section{Related Works}
An object detector is commonly trained with standard cross-entropy loss for classifying the category of each bounding box and robust loss for regressing the size of the detection box~\cite{ren2015faster,liu2016ssd,girshick2015fast,redmon2016yolo,sermanet2014overfeat}.
Some recent work has, however, begun to address modifications of the loss function when training an object detector in order to improve performance.
Shrivastava~\etal~\cite{shrivastava2016training} proposed online hard example mining (OHEM) that only backpropagates losses of hard examples, thus making the network focus on discriminating difficult cases.
Focal loss~\cite{lin2018focal} is another method that proposes to attenuate the loss the more confident the network is about a prediction, which also leads to a similar effect as OHEM.
Contrary to these methods, our proposed part-aware sampling fully ignores losses of probable unannotated false negatives during training. Loss attenuation methods can be applied jointly with our method, making our approach orthogonal to these previous works. 

Prediction results of a network are used to train an object detector with limited annotation in many previous works~\cite{shi2012weakly,siva2011weakly,bilen2016weakly,gokberk2016weakly,tang2014weakly}.
Yan~\etal~\cite{yan2017ssl} uses pseudo labels to tackle the problem where a subset of a dataset is annotated with bounding boxes and the rest of the dataset without annotations.
Inoue~\etal~\cite{inoue2018weakly} works on learning an object detector on a domain with no bounding box annotation by using pseudo labels generated by a model trained on another domain with shared categories.
Wu~\etal~\cite{wu2018softsampling} proposes score-based soft sampling, which uses pseudo labels to complement sparse annotations.
The method is studied only in PASCAL VOC, which is quite small in today's standard.
Although they conclude the technique to be unreliable, their conclusion assumes that performance of an object detector is weak, which may not be true in the case when a large network is trained on a large dataset.


The work of Zhang~\etal~\cite{zhang2018part} targets the task of fine-grained object classification.
In their method, they train a weakly supervised network for detecting discriminative local parts that can be used for fine-grained classification. Their approach is similar to ours in that it is part-aware,
although it differs in the fact that while they learn a network to detect parts as an auxiliary task during training, we leverage the relationship between parts in order to prevent false negative due to missing ground truths.

Fang~\etal~\cite{fang2017object} propose a framework for training an object detector that utilizes
an external knowledge graph, exploiting knowledge about what types of objects commonly occur together.
While their approach aims to learn knowledge possibly not in the training set, such as that a cat often sits on a table, our approach directly aims to tackle the issue of missing ground truth annotations in an image, which deteriorates object detection performance by increasing false negatives.


\section{Problems}
Open Images Dataset v4 (OID) is a recently introduced object detection dataset on an unprecedented scale in terms of the numbers of annotated images and bounding boxes as well as the number of categories supported.
The dataset is different from its predecessors not only in terms of its size but also in terms of its requirement on an annotation by allowing a subset of categories to be not annotated even if the categories are present in an image.
For each image, annotation only covers a set of categories called \textit{verified categories}, which is a subset of categories that annotators check for existence in an image.
In some images, verified categories do not span all categories present in an image, resulting in a subset of present instances not annotated.
Verified categories consist of positively verified and negatively verified categories.
Positively verified categories exist in an image, and negatively verified categories are checked by human annotators not to exist in the image.
Figure~\ref{fig:sparse_verified} demonstrates two example images containing people with different verified categories.
Since a much larger set of human part categories are verified in the bottom image, the annotations of these categories are denser.

In OID, there are on average $7.4$ categories in verified categories for each image, which is much fewer than the $500$ categories supported by the dataset.
Moreover, although the number of supported categories are more than six times larger than COCO, OID contains on average almost the same number of positively verified categories (i.e., $3.4$ and $2.9$ categories per image for OID and COCO, respectively) implying that the annotations of OID are much more sparse than COCO.
In fact, the authors of OID reported that the recall of positively verified categories is only $43\%$~\cite{kuznetsova2018openimages}.

\section{Methods}
We explore two methods of determining objects that are unannotated in a sparsely annotated dataset.
First, we propose \textit{part-aware} sampling, which ignores classification loss for part categories when an instance of them is inside an instance of their subject categories.
Second, we use pseudo labels generated from a pretrained model to exclude regions that are likely not to be annotated.
Despite the idea of pseudo labels being widely recognized~\cite{yan2017ssl,wu2018softsampling}, there is not much consensus on how to utilize it, especially for object detection.
We propose a pipeline to filter unreliable pseudo labels using cues that are available from the structure of the problem.
We call the method \textit{pseudo label-guided} sampling.

\subsection{Basic Architecture}
We use a proposal-based object detector like Faster R-CNN~\cite{ren2015faster} in this work.
The detection pipeline consists of a region proposal network that produces a set of class-agnostic region proposals around instances and the main network that classifies each proposal into categories and refines it to better localize an instance.
A proposal is classified to the background category or one of the foreground categories.
During training, a category is assigned to each proposal, and this assignment is used to calculate classification loss and localization loss like Faster R-CNN~\cite{ren2015faster}.
In this work, the classification loss is calculated as the sum of sigmoid cross entropy loss for each proposal and each category as:
\begin{equation}
\begin{aligned}
\mathcal{L}_{cls} &= - \sum_i \sum_c l_{ic}\log p_{ic} \\
l_{ic} &\in \{-1, 0, 1\}~,
\end{aligned}
\end{equation}
where $l_{ic} = 1$ and $l_{ic} = -1$ when the $i$-th proposal is assigned or not assigned to category $c$, respectively.
Also, $l_{ic}$ can be set to $0$, which means that the classification loss for category $c$ is ignored for the $i$-th proposal.
We explore later in this section how to determine ignored categories for each RoI proposal, which plays a critical role in diminishing incorrect training signals created by missing annotations.

\subsection{Part-Aware Sampling}
For certain pairs of categories, one is a part or a possession of the other in most of the images where they co-occur.
We call the categories in such kind of a pair as \emph{part category} and \emph{subject category}.
For instance, human parts like faces are usually parts of people, and tires are often parts of cars.
Also, accessories and clothes are in many cases possessed by people in the images of the OID dataset.
Furthermore, for these pairs of categories, we find that annotation is often lacking for part categories.

In Table~\ref{tab:dataset_cooccur_stat}, we show statistics that supports our observation of part and subject relationships.
First, we measure the ratio of a bounding box of a part category to be included in a subject category as shown in row \textit{included}.
We determine if a box $b_1$ is included in another box $b_2$ when asymmetric intersection over union ($aiou(b_1, b_2) = \frac{area(b_1 \cap b_2)}{area(b_1)}$) is higher than a certain threshold $\tau$.
The ratio \textit{included} is formally computed as

\begin{equation}
\begin{aligned}
	\frac{\# \{b_p \in \mathcal{B}_p \mid \exists b_s \in \mathcal{B}_s, aiou(b_p, b_s) > \tau \}}{\#\mathcal{B}_p}~,
\end{aligned}
\end{equation}
where $\mathcal{B}_p$ and $\mathcal{B}_s$ are the sets of bounding boxes of a part category $p$ and a category $s$, which is the subject category of $p$.
Note that we only consider bounding boxes in a set of images $\mathcal{I}_p \cap \mathcal{I}_s$, where $\mathcal{I}_c$ is the set of images that contain category $c$.
Furthermore, in row \textit{co-occur} of Table~\ref{tab:dataset_cooccur_stat}, we show the ratio of images that contain annotations of both part and subject categories, which is formulated as  $\frac{\#\mathcal{I}_p \cap \mathcal{I}_s}{\#\mathcal{I}_s}$.

From row \textit{included}, We can observe that a part category is included in its subject category in more than $90\%$ of bounding boxes for $39$ out of $47$ pairs.
Thus, the part and subject relationships are reflected in spatial relationships of objects in OID.
From row \textit{co-occur}, the percentages are too small for many pairs of categories based on our common sense, implying that annotation is severely missing for part categories.
For instance, the percentage of human eyes is only $2.8\%$.

To reduce false training signals, we introduce part-aware sampling that selects categories to ignore for classification loss from RoI proposals.
The main idea behind this technique is that given the high likelihood that instances of part categories are included in subject categories, it is safer to ignore classification loss for part categories for RoI proposals included in a subject category.
The technique is used only when part categories are not included in verified categories.
Figure~\ref{fig:pas} illustrates the technique by visualizing a subset of RoI proposals that are ignored for classification loss of parts of people and cars.

In our work, we use statistics collected as in Table~\ref{tab:dataset_cooccur_stat} and prior knowledge of category relationships to design a mapping $\mathcal{P}$, which maps a label to its part categories.
Algorithm~\ref{algo:part_aware_sampling} summarizes this method.

\begin{table*}[tp]
\centering
\caption{Statistics of part and subject categories. See the text for definitions of \emph{included} and \emph{co-occur}.
} \label{tab:dataset_cooccur_stat}
\footnotesize \setlength{\tabcolsep}{2.3pt}         
\begin{tabulary}{\linewidth}{lcccccccccccccccc}\toprule[1pt]
Subject & \multicolumn{16}{c}{Person}\\
Part & Arm & Ear & Nose & Mouth & Hair & Eye & Beard & Face & Head & Foot & Leg & Hand & Glove & Hat & Dress & Fedora  \\     \midrule
Included (\%) & $91.3$ & $94.0$ & $93.1$ & $94.7$ & $89.2$ & $94.7$ & $99.2$ & $91.8$ & $88.1$ & $87.9$ & $90.5$ & $90.3$ & $95.0$ & $93.7$ & $97.5$ & $94.5$    \\
Co-occur (\%) & $4.87$ & $0.98$ & $3.40$ & $2.90$ & $6.75$ & $2.81$ & $0.30$ & $39.7$ & $5.76$ & $0.05$ & $1.98$ & $2.40$ & $0.05$ & $0.99$ & $3.77$ & $0.33$   \\

\bottomrule
\toprule
Subject & \multicolumn{16}{c}{Person}\\
Part & Footwe. & Sandal & Boot & Sports. & Coat & Sock & Glasse. & Belt & Helmet & Jeans & High h. & Scarf & Swimwe. & Earrin. & Bicycl. & Shorts \\     \midrule
Included (\%) & $84.6$ & $89.8$ & $90.8$ & $95.0$ & $92.9$ & $85.5$ & $96.6$ & $96.7$ & $93.7$ & $91.2$ & $93.0$ & $97.4$ &  $92.1$ & $96.7$ & $94.8$  & $92.2$    \\
Co-occur (\%) & $15.2$ & $0.07$ & $0.08$ & $0.07$ & $0.37$ & $0.02$ & $5.04$ & $0.02$ & $0.82$ & $3.67$ & $0.09$ & $0.20$ & $0.36$ & $0.02$ &  $0.55$ & $0.79$    \\

\bottomrule
\toprule
Subject & \multicolumn{11}{c|}{Person} & \multicolumn{3}{c|}{Car} & \multicolumn{1}{c|}{Door} \\
Part & Baseba. & Minisk. & Cowboy. & Goggles & Jacket & Shirt & Sun ha. & Suit &  Trouse. & Brassi. & Tie & Licens. & Wheel & Tire & Handle &  \\       \midrule 
Included (\%) & $96.1$ & $98.6$ & $93.3$ & $95.6$ & $94.8$ & $97.9$ & $91.8$ & $96.7$ & $90.8$ & $98.2$ & $98.5$ & $91.3$ & $80.6$ & $79.2$ & $95.3$ &   \\
Co-occur (\%) & $0.17$ & $0.01$ & $0.21$ & $0.79$ & $1.51$ & $0.59$ & $0.48$ & $5.38$ & $0.54$ & $0.11$ & $0.91$ & $2.52$ & $39.97$ & $14.5$ & $1.65$ &   \\
\bottomrule[1pt]
\end{tabulary} \vspace{-8pt} 
\end{table*} 

\begin{algorithm}[t]
    \hyphenation{object}
    \caption{Framework of \textit{Part-Aware Sampling}} 
    \hspace*{0.02in} {\bf Input:} 
    RoI proposals $\{r_i\}_{i=1}^{N}$,
    ground truth boxes $\{b_j\}_{j=1}^M$, ground truth labels $\{l_j\}_{j=1}^M$, verified labels $\mathcal{V}$,
    and a set $\mathcal{P}$ that maps a subject category to the list of its part categories. \\
    \hspace*{0.02in} {\bf Initialize:} Set $\mathcal{I} \leftarrow \{\{\}_i\}_{i=1}^N$
    \begin{algorithmic}[1]
        \For{$i=1$ to $N$}
        \For{$j=1$ to $M$}
        \If{$aiou(r_i, b_j) > \tau$ and $l_j$ in $\mathcal{P}$}
        \For{$p$ in $\mathcal{P}[l_j]$}
        \If{$p$ not in $\mathcal{V}$}
        \State{Append $p$ to $\mathcal{I}_i$}
        \EndIf
        \EndFor
        \EndIf
        \EndFor
        \EndFor
    \end{algorithmic}
    \hspace*{0.02in} {\bf Output:} Set of categories $\mathcal{I}$, which are ignored when calculating classification loss for each RoI proposal.
    \label{algo:part_aware_sampling}
\end{algorithm}

\subsection{Pseudo Label-Guided Sampling}
\label{subsec:pseudo-label}
Alternatively to part-aware sampling,
we also present pseudo label-guided sampling to tackle the sparse annotation problem.
For this method, we train a network twice and use the pseudo labels generated from the first model to guide the training of the second model.

We filter prediction results of a trained model to generate pseudo labels to complement sparse annotation for unannotated regions.
We first ignore all prediction results with the categories included in verified labels.
Second, we ignore predicted boxes with high IoU with any of the actual ground truths because these predicted boxes are likely the result of misclassifying an annotated instance.
Third, prediction results with scores below a score threshold are rejected.
The score threshold $\mathcal{T}$ is determined for each category based on the precision on the withheld dataset at different score thresholds.
The minimum precision is specified as a hyperparameter that is used to determine the score threshold $\mathcal{T}$ by setting it as the minimum threshold that achieves the precision.

\begin{figure}
\subfloat[
]{
	\label{subfig:pseudo_gt_0}
	\includegraphics[width=0.15\textwidth]{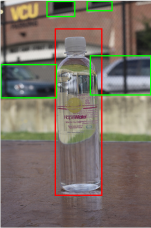} }
\hfill
\subfloat[
]{
	\label{subfig:pseudo_gt_1}
	\includegraphics[width=0.3\textwidth]{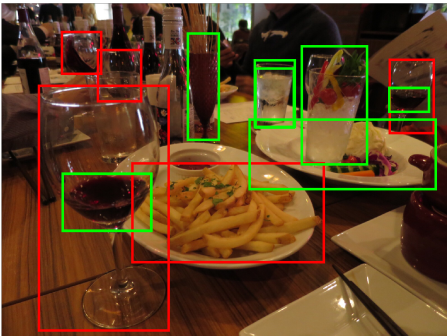} } 
\caption{Examples of pseudo labels.
Red bounding boxes are the ground truths annotations and green bounding boxes are pseudo labels. (a): "Bottle" is annotated. "Windows" and "cars" are included in the pseudo labels. (b): "French fry" and "wine glass" are annotated. "Wine", "cocktail" and "plate" are included in the pseudo labels.}
\label{fig:pseudo_gt_example}
\end{figure}

The algorithm is summarized in Algorithm~\ref{algo:pseudo_gt}.
Figure~\ref{fig:pseudo_gt_example} shows examples of pseudo labels generated by the algorithm.

\begin{algorithm}[t]
    \hyphenation{object}
    \caption{Framework of \textit{Pseudo Label-Guided Sampling}} 
    \hspace*{0.02in} {\bf Input:} 
    RoI proposals $\{r_i\}_{i=1}^{N}$,
    ground truth boxes $\{b_j\}_{j=1}^M$, ground truth labels $\{l_j\}_{j=1}^M$, and verified labels $\mathcal{V}$.
    Also, output of a pre-trained model that includes bounding boxes $\{\hat{b_k}\}_{k=1}^L$, labels $\{\hat{l_k}\}_{k=1}^L$, and scores $\{\hat{s_k}\}_{k=1}^L$.
    Also, $\mathcal{T}$ is a set of score thresholds for each category. \\
    \hspace*{0.02in} {\bf Initialize:} Set $\mathcal{I} \leftarrow \{\{\}_i\}_{i=1}^N$
    \begin{algorithmic}[1]
        \State{$\mathcal{K} = \{1, \cdots, L\}$}
        \For{$k=1$ to $L$}
        \If{$\hat{s}_k < \mathcal{T}[\hat{l}_k]$ or $\hat{l}_k$ in $\mathcal{V}$}
        \State{Remove $k$ from $\mathcal{K}$; Continue}
        \EndIf
        \For{$j=1$ to $M$}
        \If{$iou(b_j, \hat{b}_k) > 0.8$ }
        \State{Remove $k$ from $\mathcal{K}$; Break}
        \EndIf
        \EndFor
        \EndFor

        \For{$i=1$ to $N$}
        \For{$k$ in $\mathcal{K}$}
        \If{$iou(b_i, \hat{b}_k) > 0.5$}
        \State{Append $\hat{l}_k$ to $\mathcal{I}_i$}
        \EndIf
        \EndFor
        \EndFor
    \end{algorithmic}
    \hspace*{0.02in} {\bf Output:} Set of categories $\mathcal{I}$, which are ignored when calculating classification loss for each RoI proposal.
    \label{algo:pseudo_gt}
\end{algorithm}

\section{Experiments}
We conduct experiments on sparse COCO and Open Images Dataset v4 (OID).
Sparse COCO is a dataset created from MS COCO~\cite{yi2014coco} that contains sparse annotation.
Since the size of sparse COCO is much smaller than OID and has access to complete annotations for all objects for analysis, we use this dataset to study  pseudo label-guided sampling and the negative effect of missing annotations in general before experimenting on OID.
Since part and subject relationships are not common in sparse COCO, we were only able to experiment part-aware sampling with OID.

\subsection{Implementation Details} 
We use Feature Pyramid Networks~\cite{lin2017feature} for our experiments.
The feature extractor is ResNet50~\cite{he2016deep} for experiments using sparse COCO and SE-ResNeXt50~\cite{hu2017squeeze} for experiments using OID.
The larger network is selected for Open Images Dataset because the capacity of the base extractor needs to be large enough to learn such a large dataset.
The initial bias for the final classification layer is set to a large negative number to prevent the training from getting unstable in the beginning.
We set the initial weight of the base extractor with the weights of an image classification network trained on the ImageNet classification task~\cite{imagenet_cvpr09}.
We use stochastic gradient descent with momentum set to $0.9$ for optimization.
The base learning rate is set to $0.00125 \times \text{batchsize}$.
We use a warm-up learning rate schedule to stabilize training in the beginning.
For sparse MS COCO, we trained for $90000$ iterations with 16 images in each batch. The learning rate is multiplied by $0.1$ at the $60000$-th and the $80000$-th iterations.
For OID, we trained for $12$ epochs.
The learning rate is scheduled by a cosine function $\eta = \eta_0 \frac{\cos{(\text{\% of progress} \times \pi)} + 1}{2}$, where $\eta$ and $\eta_0$ are the learning rate and the initial learning rate.
We scale images during training so that the length of the smaller edge is between $[650, 1056]$.
Also, we randomly flip images horizontally to augment training data.
We use Chainer~\cite{tokui2015chainer,akiba2017chainermn,niitani2017chainercv} as our deep learning framework.

\subsection{Sparse COCO}
Sparse COCO is a dataset artificially created by randomly deleting labels in images of MS COCO.
For each category in MS COCO, among the set of images containing the category,
we delete all annotations of the category for the images selected from the set by probability $\alpha$.
This means that for each image in the artificially created dataset,
the instances of the labeled categories are annotated exhaustively as in the original MS COCO dataset, but there could be categories with no annotation even if instances of the categories exist.
Table~\ref{tab:coco_statistics} shows the statistics of the dataset created artificially with different probabilities (0.3, 0.5, 0.7).

\begin{table}[t]
\centering \addtolength{\tabcolsep}{-2pt}
\footnotesize
\caption{Statistics of sparse COCO for different probabilities $\alpha$ of deleting annotations.
}\label{tab:coco_statistics}
\begin{tabular}{l|cccc}
\toprule
 & 0 &  0.3 & 0.5 & 0.7  \\
\midrule
number of boxes per image & 2.90  &  5.03 & 3.60   & 2.17 \\
number of distinct categories per image & 7.19 &  2.03  & 1.45 & 0.87 \\
\bottomrule
\end{tabular}\vspace{0.1cm}
\vspace{-0.03in}
\vspace{-0.25cm}
\end{table}

We use the training set of COCO 2017 object detection challenge to create sparse COCO for training networks and tuning hyper parameters.
The validation split is used without deleting annotations for validation.
We evaluate models using mmAP used in the COCO competition.

We first evaluate different methods with different level of missing annotations.
Among the methods we tried, we fixed every setting except the way different RoI proposals are evaluated as positive, negative or ignored samples for training the classification network.
Note that we do not compare with part-aware sampling since COCO categories do not include part categories.
Here are the methods:
\begin{itemize}
\item \textbf{Baseline:} This method follows the standard training procedure that assumes a dataset with an exhaustive annotation.
RoIs around instances that are not annotated are evaluated falsely as negative samples for this method.
\item \textbf{Oracle ignore:} This method uses the ground-truth that are deleted in order to evaluate how much performance loss can be recovered by labeling RoIs using oracle information.
For any RoI proposals that have IoU with the deleted ground-truth higher than $0.5$, this method ignores classification loss calculated from them.
\item \textbf{Oracle positive:} Similarly to "oracle ignore" described above, this method also uses the ground-truth that is deleted.
Instead of ignoring RoIs overlapping with the deleted ground-truth during training, this method uses those RoIs as positive samples. The difference between this method and training on a fully annotated dataset is that during the sampling of RoIs that are actually used for training, this method does not use the deleted ground-truth to oversample regions around the ground-truth, thus making the comparison with other methods fair.
\item \textbf{Pseudo label-guided sampling:} The method is described in Section~\ref{subsec:pseudo-label}.
Score thresholds are selected by training another model with the default training scheme using $80\%$ of the training data
and using the remaining $20\%$ of the training data to calculate precisions at different score thresholds.
We also experimented assigning positive labels to RoIs that overlap with highly confident pseudo labels.
\item \textbf{Overlap-based soft sampling~\cite{wu2018softsampling}:} This method multiplies weights on the loss computed using negative samples.
The weights are determined based on overlaps with annotated bounding boxes.
The method is designed based on the assumption that regions close to annotated ground truth can be confidently assigned to the background.
The method is demonstrated to work well with PASCAL VOC according to the authors.
We use the same values for all hyper parameters of the nonlinear function that takes overlap as input and weight multiplied on the loss as output.
Since the scale of the total loss changes from the rest of the methods, for a fair comparison, we choose the optimal learning rate by searching over a set of learning rates that are $\times \frac{1}{2}, \times 1, \times 2, \times 3, \times 4$ of the learning rate used by the rest of the methods.
\end{itemize}

Table~\ref{tab:partial_coco_results} summarizes the main result from sparse COCO.
The model trained using full annotation obtains 36.75 mmAP on the validation set, and this can be considered as the maximum score that any methods can obtain.
We have the following observations.
First, the performance recovers from the baseline by using oracle information to ignore proposals.
The amount of negative effect caused by missing annotation increases as the ratio of missing annotation increases.
The difference between \textit{Baseline} and \textit{Oracle ignore} is $0.70$ mmAP and $1.67$ mmAP when $\alpha=0.3$ and $\alpha=0.7$, respectively.

Second, by using pseudo-ground truths to decide instances that are falsely annotated as negatives, the result matches the method using oracle information to ignore samples.
The pseudo labels cover regions included in oracle information and also regions that a network falsely detected.
The result suggests that training may work better when ignoring regions that are unannotated, but susceptible to mistakingly recognizing as the foreground.

Third, despite making extra efforts to tune parameters, overlap-based soft sampling~\cite{wu2018softsampling} performs worse than the baseline on sparse COCO.
Unlike the other methods, this method discourages contribution of negative samples unanimously based on their distance from the closest ground truth bounding box.
Perhaps, this method discourages too many negative samples from contribution to the loss for this dataset.

Table~\ref{tab:ablative_score_threshold} shows an ablative study of methods using pseudo labels.
We make the following observations.
First, performance improves by selecting score thresholds for each category based on category-wise precision compared to uniformly setting the values.
Second, the performance is relatively robust to different precision thresholds, but the precision threshold at $0.5$ works the best.
Third, performance drops by using pseudo labels to assign positive labels to RoI proposals.
This is contrary to our expectation that a network learns better by assigning proposals to positives instead of ignoring them when pseudo-labels are created highly confident prediction results.
We think that the object detector is not robust to false positives because the false positives play a big role due to the number of positive samples being small.

\begin{table}
\centering \addtolength{\tabcolsep}{-2pt}
\footnotesize
\caption{
Comparison of different methods on sparse MS COCO.
A model trained on COCO with complete annotation achieves
36.75 mmAP.}\label{tab:partial_coco_results}
\begin{tabular}{l|ccc}
\toprule
 & 0.3 & 0.5 & 0.7  \\
\midrule
Baseline &  34.22  & 31.69  & 27.31 \\
Oracle Ignore & 34.92 &  32.73 & 28.98 \\
Pseudo Label-guided (Ours) & 35.00 & 32.79 & 29.03  \\
Overlap-based soft~\cite{wu2018softsampling} & 33.98 & 31.39 & 27.30  \\
Oracle Positive & 35.66 & 34.17 & 32.19  \\
\bottomrule
\end{tabular}\vspace{0.1cm}
\vspace{-0.03in}
\end{table}

\begin{table}
\centering \addtolength{\tabcolsep}{-2pt}
\footnotesize
\caption{Comparison of different score thresholds for pseudo label-guided sampling.
If pseudo labels are not used to select positive RoI proposal samples, the second column is left empty.
Uniform ($x$) indicates that the constant threshold $x$ is used for all categories.
Prec ($>y$) indicates that thresholds are selected for each class differently based on the minimum tolerable precision $y$.
}
\label{tab:ablative_score_threshold}
\begin{tabular}{cc|c}
\toprule
ignore threshold & positive threshold  & mmAP \\
\midrule
uniform ($0.3$) &  &  34.76  \\
uniform ($0.5$) &  &  34.71  \\
prec($>0.3$) &  &  34.94   \\
prec($>0.5$) &  &  35.00   \\
prec($>0.7$) &  &  34.93   \\
prec($>0.5$) & prec($>0.8$)  & 34.59   \\
\bottomrule
\end{tabular}\vspace{0.1cm}
\vspace{-0.25cm}
\end{table}

\subsection{Open Images Dataset v4}
Open Images Dataset v4 (OID) is a newly introduced dataset that can be used for object detection.
We use the split of the data and the subset of the categories that were used for the competition held in 2018 hosted by dataset authors.~\footnote{\url{https://storage.googleapis.com/openimages/web/challenge.html}}
The training and the validation split contain $1,643,042$ and $100,000$ images, respectively.
There are $500$ distinct categories annotated with bounding boxes in OID.
These categories have clearly defined spatial extents and considered as important concepts by the dataset authors.

Table~\ref{tab:oic_results} summarizes the main results using OID.
The baseline follows the standard training procedure and does not use any special technique to tackle missing annotation.
We use the precision threshold at $0.5$ for pseudo label-guided sampling.
Both pseudo label-guided sampling and part-aware sampling improve upon the baseline.
Although pseudo label-guided sampling performs competitively even on results using oracle information for sparse COCO, part-ware sampling achieves better results on OID.
Since the ratio of missing annotation is sometimes lower than $10\%$ as suggested from Table~\ref{tab:dataset_cooccur_stat}, we suggest that it is difficult to train a pretrained model for some categories in OID.

In Table~\ref{tab:cooccur_class_score}, we take a closer look at evaluation results by examining category-wise AP for part categories.
The part-aware sampling leads to on average $0.7$ AP improvement across all categories and $9.2$ AP improvement on part categories.
In particular, for human part categories, such as "human face" and "human ear", we see a significant improvement of $22.7$ AP on average.

\begin{figure}
\subfloat[
]{
	\label{subfig:qualitative_0_0}
	\includegraphics[width=0.22\textwidth]{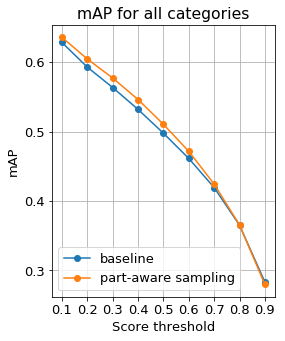} }
\subfloat[
]{
	\label{subfig:qualitative_1_0}
	\includegraphics[width=0.22\textwidth]{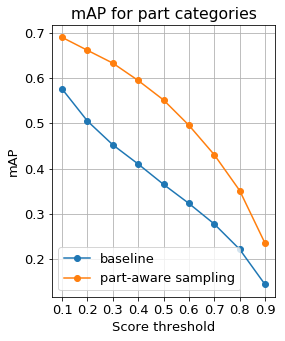} }
\\
\caption{mAP on OID at different score thresholds for the baseline and part-aware sampling.}
\label{fig:score_map_figure}
\end{figure}

Figure~\ref{fig:score_map_figure} shows the averages of APs at different score thresholds for all categories and the subset of categories that are used as part categories.
The difference between the baseline and part-aware sampling is already large for part categories with low score threshold, but the gap widens as the score threshold increases.

\begin{figure}
	\includegraphics[width=0.48\textwidth]{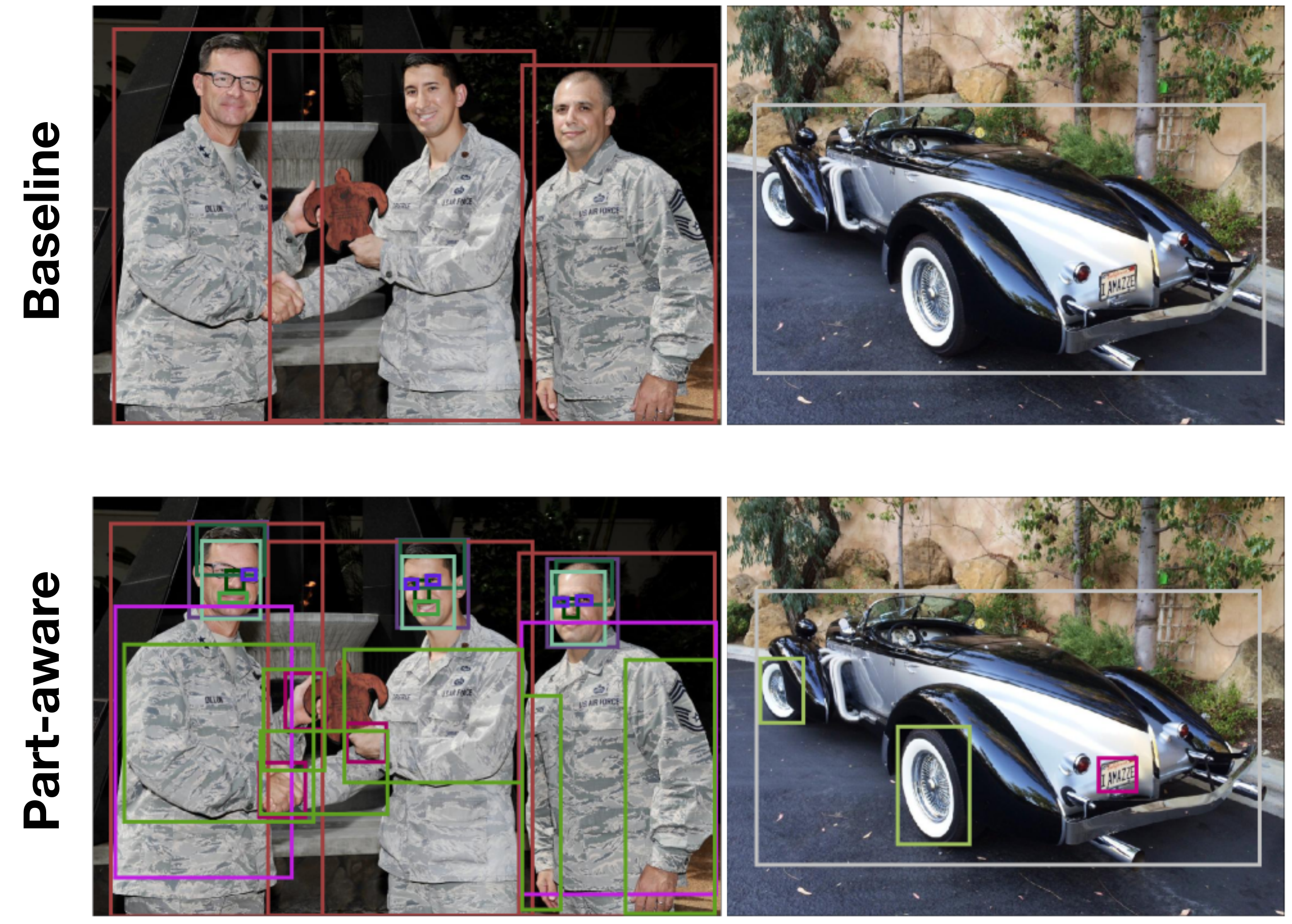} 
\caption{The visualization of outputs of models trained without part-aware sampling (top-row) and with it (bottom-row) on OID. The score thresholds are kept the same for all images.
}
\label{fig:qualitative_comparison}
\end{figure}

In Figure~\ref{fig:qualitative_comparison} shows a qualitative comparison of models trained with and without part-aware sampling.
For the models with part-aware sampling, part categories are detected with a relatively high score threshold.
For instance, in the right image, tires and license plates are only detected by the model trained with part-aware sampling.

\paragraph{Open Images Competition 2018:}
Based on the model trained with part-aware sampling, we integrate context head~\cite{zhu2017couplenet}, longer training time, a stronger feature extractor~\cite{hu2017squeeze}, an additional number of anchors, and test-time augmentation for our submission to the object detection track of Google AI Open Images Competition 2018.
For evaluation, the test set is split into the public and private sets.
During the period of the competition, scores on the public set were always available to the competitors, but scores on the private set were not disclosed until the end of the competition.
Our best single model achieves $55.81$ mAP and $53.43$ mAP on public and private sets.
Our ensemble of models achieves $1$st and $2$nd best scores on the public and private sets with $62.88$ mAP and $58.63$ mAP.
Table~\ref{tab:competition} summarizes the result of ours and other top competitors.

\begin{table}[t]
\centering \addtolength{\tabcolsep}{-2pt}
\footnotesize
\caption{Results on the validation set of Open Images Dataset v4.}\label{tab:oic_results}
\begin{tabular}{l|c}
\toprule
 & validation mAP    \\
\midrule
Baseline &  $64.49$  \\
Pseudo label-guided sampling   & $64.84$  \\
Part-aware sampling & $\mathbf{65.18}$   \\
\bottomrule
\end{tabular}\vspace{0.1cm}
\vspace{-0.03in}
\vspace{-0.25cm}
\end{table}
\begin{table*}[htp]
\centering
\caption{Ablative study of part-aware sampling on categories that can be ignored by the technique. The scores are AP calculated on the validation set of OID.} \label{tab:cooccur_class_score}
\footnotesize \setlength{\tabcolsep}{2.3pt}         
\begin{tabulary}{\linewidth}{lcccccccccccccccc}\toprule[1pt]

& Arm & Ear & Nose & Mouth & Hair & Eye & Beard & Face & Head & Foot & Leg & Hand & Glove & Hat & Dress & Fedora  \\     \midrule

Baseline & $40.9$ & $17.5$ & $34.7$ & $21.4$ & $63.8$ & $27.3$ & $55.5$ & $82.7$ & $55.1$ & $50.7$ & $41.6$ & $32.3$ & $\mathbf{63.4}$ & $64.9$ & $70.6$ & $67.0$   \\
Part-aware sampling & $\mathbf{55.2}$ & $\mathbf{62.6}$ & $\mathbf{69.6}$ & $\mathbf{55.2}$ & $\mathbf{74.7}$ & $\mathbf{64.0}$ & $\mathbf{76.8}$ & $\mathbf{91.4}$ & $\mathbf{78.9}$ & $\mathbf{59.5}$ & $\mathbf{54.4}$ & $\mathbf{53.6}$ & $60.8$ & $\mathbf{69.0}$ & $\mathbf{73.9}$ & $\mathbf{70.3}$   \\

\bottomrule
\toprule
& Footwe. & Sandal & Boot & Sports. & Coat & Sock & Glasse. & Belt & Helmet & Jeans & High h. & Scarf & Swimwe. & Earrin. & Bicycl. & Shorts \\     \midrule
Baseline & $61.9$ & $53.6$ & $\mathbf{61.6}$ & $52.9$ & $58.0$ & $\mathbf{70.6}$ & $74.9$ & $\mathbf{66.8}$ & $80.2$ & $62.7$ & $76.6$ & $71.6$ & $\mathbf{63.4}$ & $82.0$ & $75.1$ & $69.7$   \\
Part-aware sampling & $\mathbf{68.5}$ & $\mathbf{58.9}$ & $57.9$ & $\mathbf{61.2}$ & $\mathbf{73.3}$ & $67.1$ & $\mathbf{85.4}$ & $61.9$ & $\mathbf{82.4}$ & $\mathbf{77.6}$ & $\mathbf{78.8}$ & $\mathbf{75.8}$ & $\mathbf{63.4}$ & $\mathbf{86.1}$ & $\mathbf{75.8}$ & $\mathbf{75.4}$  \\

\bottomrule
\toprule
& Baseba. & Minisk. & Cowboy. & Goggles & Jacket & Shirt & Sun ha. & Suit &  Trouse. & Brassi. & Tie & Licens. & Wheel & Tire & Handle & \underline{Average} \\       \midrule 
Baseline & $\mathbf{67.2}$ & $\mathbf{62.5}$ & $65.0$ & $79.3$ & $69.5$ & $70.9$ & $61.3$ & $83.7$ & $62.5$ & $\mathbf{82.6}$ & $84.7$ & $72.1$ & $48.3$ & $49.4$ & $41.1$ & $61.1$   \\
Part-aware sampling & $62.2$ & $58.7$ & $\mathbf{73.3}$ & $\mathbf{86.7}$ & $\mathbf{74.3}$ & $\mathbf{81.6}$ & $\mathbf{66.4}$ & $\mathbf{87.0}$ & $\mathbf{69.8}$ & $74.5$ & $\mathbf{91.5}$ & $\mathbf{74.6}$ & $\mathbf{66.4}$ & $\mathbf{69.6}$ & $\mathbf{46.2}$ & $\mathbf{70.3}$  \\
\bottomrule[1pt]
\end{tabulary} \vspace{-8pt} 
\end{table*} 
\begin{table}
\centering \addtolength{\tabcolsep}{-2pt}
\footnotesize
\caption{Results on the test set of OID. Unlike the other results, test-time augmentation is used.}\label{tab:competition}
\begin{tabular}{l|ccc}
\toprule
 & public test & private test  \\
\midrule
Single best (Ours) &  $55.81$ & $53.43$ \\
Ensemble (Ours)  &  $\mathbf{62.88}$  & $\underline{58.63}$ \\
Private LB 1st place &  $61.71$ & $\mathbf{58.66}$ \\
Private LB 3rd place~\cite{gao2018baidu} & $\underline{62.16}$ & $58.62$ \\
\bottomrule
\end{tabular}\vspace{0.1cm}
\vspace{-0.03in}
\end{table}

\section{Discussions and Future works}
In this paper, we proposed part-aware sampling and pseudo label-guided sampling to train object detectors on datasets with sparse annotation.
On Open Images Dataset v4, our part-aware sampling significantly improved results over the baseline for \textit{part categories}.
The success of our method suggests the importance of choosing a 
right measure to determine the presence of unverified objects.

Indeed, our study provides no guarantee that our method is the best method for this purpose.
Trivially, if one can prepare a \textit{perfect} pretrained model that can detect the presence of unverified objects with 100\% accuracy, the method based on such model will perform optimally. 
However, the presence of such a model completely defeats the purpose for training the model, and we need to seek methods that work in a more realistic situation.
To understand the problem better, we made an extensive empirical study of detecting unannotated objects using a pretrained model that can actually be obtained on large-scale datasets~\cite{yi2014coco,kuznetsova2018openimages}.
For sparse COCO, we found the method to work very well outperforming preexisting methods~\cite{wu2018softsampling} and matching methods with access to actual ground truths.
The method, however, underperformed against part-ware sampling for OID.
This shows that a method based on a simple prior performs better when it is difficult to obtain a reliable detector.
It is our hope that our study will instigate further exploration for the method of detecting the presence of unverified objects.   

\begin{figure*}
\captionsetup[subfigure]{labelformat=empty} 
	\includegraphics[width=\textwidth]{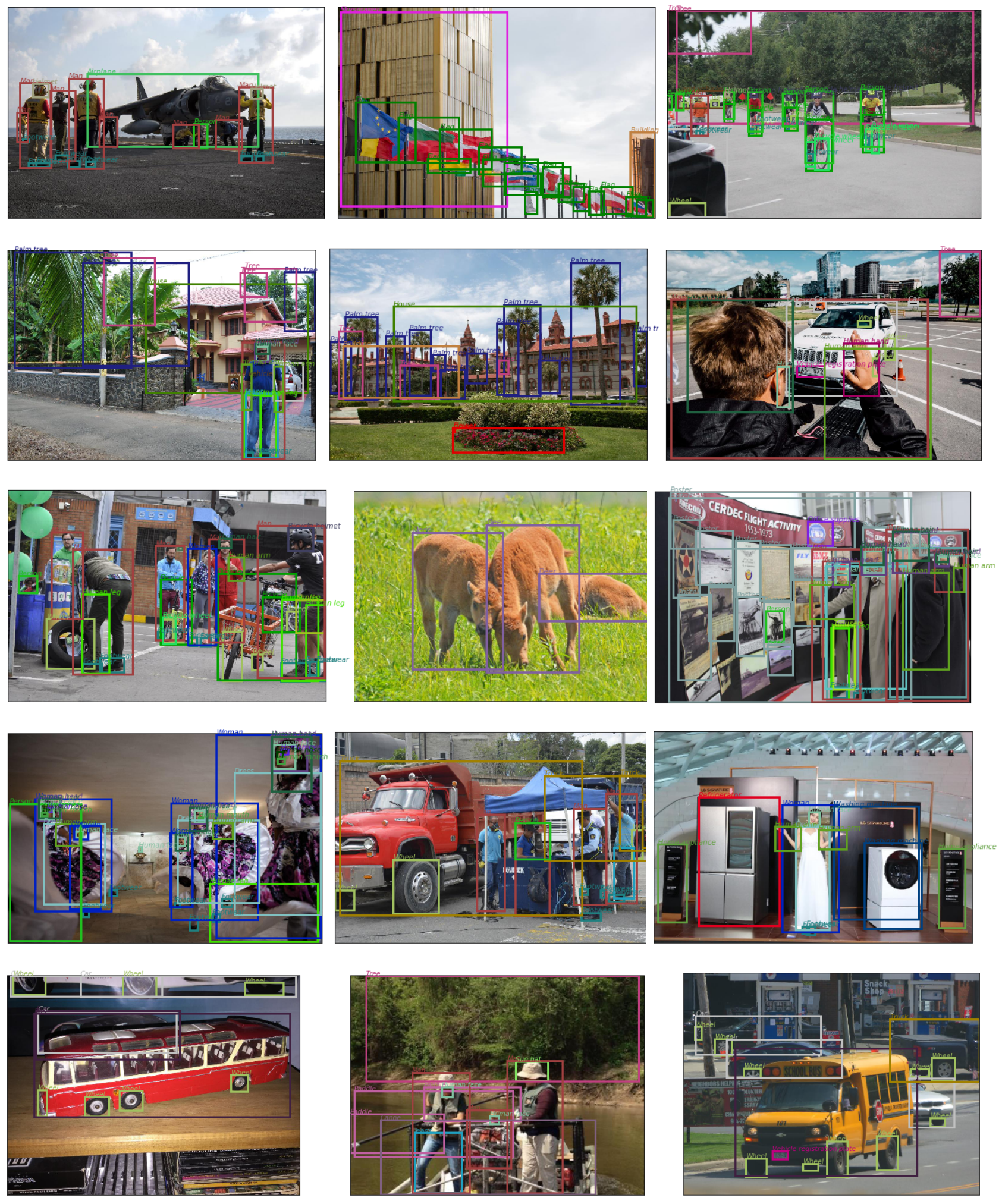}
\caption{Visualization of our trained model for the images in the test set of Open Images Dataset v4.
The best single model included in our submission to Google AI Open Images Competition 2018 is used for this visualization.
We set the score threshold to $0.5$. }
\label{fig:sup_mat}
\end{figure*}

\vspace{2em}
{
\small
\paragraph{Acknowledgments}
We thank M. Koyama for helpful insights to improve the manuscript,
and  K.~Fukuda, K.~Uenishi, R.~Arai, S.~Omura, R.~Okuta, and T.~Abe for helping the experiments.
}
\vspace{2em}

{\noindent \Large \bf Appendix}
\appendix
\section{More Results for the Competition}
We give detailed experimental results of our submission to Google AI Open Images Competition 2018.
For the competition, we added various techniques on top of \textit{part-aware sampling}. 
The improvements made by each technique is found in Table~\ref{tab:ablative_single}.

In Table~\ref{tab:ablative_ensemble}, we show the results of our single best model and ensemble of models.
The best ensemble of models includes models fine-tuned exclusively on rare categories.
These expert models improves the scores for rare categories because the single best model performs poorly on rare categories due to huge class imbalance in the dataset.
During ensembling, we further boost performance by prioritizing certain models based on their validation scores so that outputs of weaker networks do not degrade the ensemble of prediction.
We do not show the result on the validation set with this technique because the validation ground truth is used to tune parameters.
For the results of the other top competitors, we only have the scores on the test set.
We visualize a sample of our model's detections in Figure~\ref{fig:sup_mat}.

\begin{table}[H]
\centering \addtolength{\tabcolsep}{-2pt}
\footnotesize
\caption{Performance of a single model with single scale testing on the validation split with bells and whistles.}\label{tab:ablative_single}
\begin{tabular}{l|c}
\toprule
 & validation mAP  \\
\midrule
Baseline & 64.5\\
+ Part-aware sampling & 65.2 (+0.7) \\
+ 16 epochs & 65.8 (+0.6) \\
+ Context head \cite{zhu2017couplenet} & 66.0 (+0.2) \\
+ SENet-154 and additional anchors & 67.5 (+1.5)  \\
\bottomrule
\end{tabular}\vspace{0.1cm}
\vspace{-0.03in}
\end{table}

\begin{table}[H]
\centering \addtolength{\tabcolsep}{-2pt}
\footnotesize
\caption{Ensemble of models with test-time augmentation. The validation score for the other competitors' methods are not available.}\label{tab:ablative_ensemble}
\begin{tabular}{l|ccc}
\toprule
 & val & public test & private test  \\
\midrule
Single best (Ours) & 69.95 & 55.81 & 53.43 \\
Ensemble best w/o val tuning (Ours) & 74.07 & 62.34  & 58.48 \\
Ensemble best (Ours) &  & \textbf{62.88}  & \underline{58.63} \\
\midrule
Private LB 1st place &  & \underline{61.71} & \textbf{58.66} \\
Private LB 3rd place~\cite{gao2018baidu} &  & 62.16 & 58.62 \\
\bottomrule
\end{tabular}\vspace{0.1cm}
\vspace{-0.03in}
\end{table}

\newpage
{\small
\bibliographystyle{ieee}
\bibliography{egpaper_final}
}

\newpage

\end{document}